\pdfoutput=1
\newcommand{\DatasetName}{\textsc{NSF-SciFy}}
\newcommand{\DatasetNameTwentyK}{\textsc{NSF-SciFy-20K}}
\newcommand{\DatasetNameMatSci}{\textsc{NSF-SciFy-MatSci}}
\newcommand{\NSF}{National Science Foundation}
\documentclass[11pt]{article}

\usepackage[table]{xcolor}
\usepackage[final]{acl}

\usepackage{times}
\usepackage{latexsym}
\usepackage{hyperref}
\usepackage[T1]{fontenc}

\usepackage[utf8]{inputenc}

\usepackage{microtype}

\usepackage{inconsolata}

\usepackage{graphicx}

\definecolor{darkgreen}{rgb}{0,0.5,0}
\usepackage{enumitem}
\usepackage{todonotes}
\usepackage{booktabs}
\usepackage{sparklines}
\usepackage{mdframed}
\usepackage{listings}
\usepackage{float}
\usepackage{amsmath}

\usepackage[most]{tcolorbox}
\tcbset{
  samplebox/.style={
    colback=white,
    colframe=black,
    boxrule=0.5pt,
    left=1em,
    right=1em,
    top=1em,
    bottom=1em,
    arc=2pt,
    fonttitle=\bfseries
  }
}

\usepackage{multirow}

\title{\DatasetName: Mining the NSF Awards Database for Scientific Claims}

\author{Delip Rao\thanks{Corresponding author, $^\dagger$co-first author}$^\dagger$, Weiqiu You$^\dagger$, Eric Wong, Chris Callison-Burch \\
	    University of Pennsylvania \\
        Philadelphia, PA, USA \\
	    {\tt \{delip, weiqiuy, exwong, ccb\}@seas.upenn.edu} }

\begin{document}
\maketitle
\begin{abstract}
We introduce \DatasetName, a comprehensive dataset of scientific claims and investigation proposals extracted from National Science Foundation award abstracts. While previous scientific claim verification datasets have been limited in size and scope, \DatasetName~represents a significant advance with 2.8 million claims from 400,000 abstracts spanning all science and mathematics disciplines. We present two focused subsets: \DatasetNameMatSci~with 114,000 claims from materials science awards, and \DatasetNameTwentyK~with 135,000 claims across five NSF directorates. Using zero-shot prompting, we develop a scalable approach for joint extraction of scientific claims and investigation proposals. We demonstrate the dataset's utility through three downstream tasks: non-technical abstract generation, claim extraction, and investigation proposal extraction. Fine-tuning language models on our dataset yields substantial improvements, with relative gains often exceeding 100\%, particularly for claim and proposal extraction tasks. Our error analysis reveals that extracted claims exhibit high precision but lower recall, suggesting opportunities for further methodological refinement. \DatasetName~enables new research directions in large-scale claim verification, scientific discovery tracking, and meta-scientific analysis\footnote{Code and data available at \url{https://github.com/darpa-scify/NSFSciFy}}.
\end{abstract}

\section{Introduction}
\begin{figure}[h]
    \centering
    \includegraphics[width=1\linewidth]{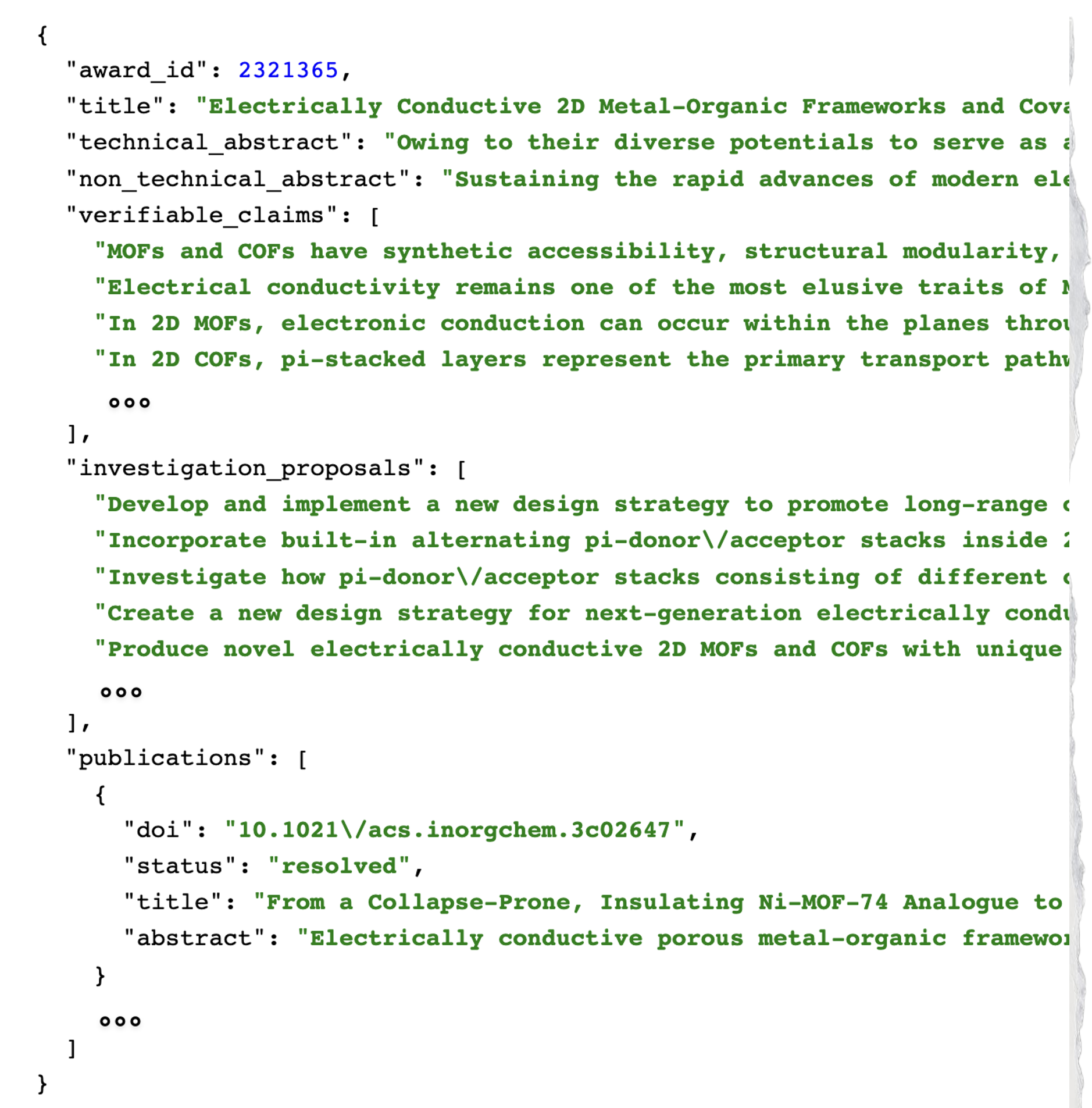}
    \caption{A sample record from our dataset. Each record contains 1) Award ID and title, 2) NSF Directorate, 3) Technical and non-technical abstracts, 4) Scientific Claims, 5) Investigation Proposals, and 6) Associated publications, when present.}
    \label{fig:sample-record}
\end{figure}
The overall growth rate of scientific publications is estimated to be 4\% annually, with a doubling time of 17 years~\cite{Bornmann2021}. Within this deluge, researchers, reviewers, and the general public struggle to separate substantiated claims from spurious ones—whether it is the ``quantum supremacy'' assertions in computing, the short-lived excitement over LK-99 superconductors\footnote[3]{for an entertaining digression c.f., \url{https://en.wikipedia.org/wiki/LK-99}}, or the misunderstanding surrounding microplastic leaches from black plastic spatulas\footnote[4]{c.f., \url{https://nationalpost.com/news/canada/black-plastic}}. Manual verification of ever growing body of scientific claims has become intractable, yet the economic and societal consequences of unverified  claims are increasingly severe. 

\begin{table*}[t]
\centering
\small
\begin{tabular}{@{}lrrll@{}}
\toprule
\textbf{Dataset} & \textbf{\# claims} & \textbf{\# docs} & \textbf{Evidence Source} & \textbf{Domain} \\
\midrule
SciFACT~\cite{wadden-etal-2020} & 1.4K & 5K & Research papers & Biomedical \\
\hline
PubHEALTH~\cite{kotonya-toni-2020}& 11.8K & 11.8K & Fact-checking sites & Public health \\
\hline
CLIMATE-FEVER~\cite{diggelmann-etal-2020} & 1.5K & 7.5K & Wikipedia articles & Climate change \\
\hline
HealthVer~\cite{sarrouti-etal-2021}& 1.8K & 738 & Research papers & Healthcare \\
\hline
COVID-Fact~\cite{saakyan-etal-2021}& 4K & 4K & Research, news & COVID \\
\hline
CoVERT~\cite{mohr-etal-2022}& 300 & 300 & Research, news & Biomedical \\
\hline
SciFACT-Open~\cite{wadden-etal-2022}& 279 & 500K & Research papers & Biomedical \\
\midrule
\textbf{\DatasetNameMatSci~(ours)} & \textbf{114K} & \textbf{16K} & \textbf{NSF award abstracts} & \textbf{Material Science} \\
\hline
\textbf{\DatasetNameTwentyK~(ours)} & \textbf{135K} & \textbf{20K} & \textbf{NSF award abstracts} & \textbf{All Science \& Math} \\
\hline
\textbf{\DatasetName~(ours)} & \textbf{2.8M} & \textbf{400K} & \textbf{NSF award abstracts} & \textbf{All Science \& Math} \\
\bottomrule
\end{tabular}
\caption{\textbf{(NSF-SciFy spans all science and math \textit{domains} and includes diverse \textit{data types}: technical/non-technical abstracts, claims, and investigation proposals.)} While previous datasets like SciFACT and PubHEALTH contain at most thousands of claims from published research papers or fact-checking sources, our \DatasetNameMatSci~and \DatasetNameTwentyK~datasets individually contribute more than 100K claims. The full \DatasetName~dataset represents an order-of-magnitude increase with  2.8M claims across 400K abstracts spanning all science \& math disciplines.  This work introduces grant abstracts as a novel, untapped source for scientific claim extraction, complementing existing approaches that focus on published literature, news articles, or social media.} 
\label{tab:datasets}
\end{table*}

\citet{wadden-etal-2020} introduced the task of scientific claim verification with the SciFACT dataset, focusing primarily on automatic verification of scientific claims. Follow up works (see Section~\ref{sec:related-work} for a detailed account) have mostly focused on the healthcare, building datasets from scientific publications, and modest-sized dataset creation. In this work, we relax all of these aspects and look at building at least an order of magnitude large-scale scientific claim dataset covering all of basic science. We envision building of such large-scale, scientific claim datasets to help future work on robust scientific claim verification systems.

We introduce \DatasetName\footnote{Short for ``NSF SCIentific FeasibilitY''.}, a comprehensive dataset of claims and investigation proposals extracted from National Science Foundation (NSF) award abstracts. We choose NSF abstracts as our source material for several reasons:

\begin{enumerate}[noitemsep,topsep=0pt]
\item NSF is a primary driver of U.S. scientific innovation, funding approximately 25\% of all federally supported basic research, spanning the entirety of science and math areas, with an annual budget of \$9.9 billion (FY 2023). Any claim dataset derived from the NSF awards database should faithfully represent the scientific Zeitgeist. 
\item NSF's rigorous subject matter expert-review process provides a high-quality filter for the claims made in funded proposals.
\item The public availability and permissive usage terms of the NSF awards database makes it an excellent resource for open science research.
\item Previous datasets on scientific claims have been derived from scientific papers, but claims in scientific grants, and particularly investigation proposals, remain unstudied.
\end{enumerate}

While not the focus of this paper, grant award abstracts additionally provide a unique opportunity to study the relationship between what researchers claim and what they propose to investigate. This could offer valuable insights into scientific practice and the evolution of research questions.

In this paper, we make the following contributions: (1) We introduce \DatasetName, the largest scientific claim dataset to date with 2.8M claims extracted from 400K NSF award abstracts, establishing grant proposals as a novel source for scientific claim extraction; (2) We create \DatasetNameMatSci~focusing exclusively on materials science with 114K extracted claims from 16K abstracts. This is the first materials science claim dataset and, in number of extracted claims, this alone is an order of magnitude bigger than the largest publicly available claim dataset; In addition, we also create \DatasetNameTwentyK~with 135K claims spanning five NSF directorates. (3) We develop a zero-shot prompting approach for joint extraction of scientific claims and investigation proposals as a scalable way to bootstrap high-precision, large-scale scientific claim datasets; (4) We present novel evaluation metrics for claim/proposal extraction based on LLM judgments, showing that fine-tuned models significantly outperform base models; and (5) Finally, we release all datasets and trained models from our work for unfettered research and commercial use. Our dataset and methods enable new opportunities for large-scale claim verification, scientific discovery tracking, and meta-scientific research.
See Appendix~\ref{app:reproduce} for reproducibility statement.

\section{Related Work}
\label{sec:related-work}

Scientific claim extraction and verification has emerged as an important research area as the volume of scientific literature continues to grow exponentially. Previous work has primarily focused on claims from published papers, fact-checking sites, and news articles.
\paragraph{Scientific Claim Datasets} Several datasets have been developed for scientific claim verification, but all have focused on claims from published literature, while we undertake the study of grant award abstracts. SciFACT \cite{wadden-etal-2020} contains 1,400 scientific claims derived from research papers in the biomedical domain. PubHEALTH \cite{kotonya-toni-2020} includes 11,800 claims from journalists and fact-checkers in public health. CLIMATE-FEVER \cite{diggelmann-etal-2020} compiled 1,500 claims from news articles about climate change. HealthVer \cite{sarrouti-etal-2021} extracted 1,800 claims from search queries related to health topics. COVID-Fact \cite{saakyan-etal-2021} and CoVERT \cite{mohr-etal-2022} focused on COVID-19 related claims from social media. SciFact-Open \cite{wadden-etal-2022} expanded the original SciFact dataset using information retrieval pooling, yet it still remains health-care focused and a few orders of magnitude smaller than our largest dataset.

\begin{figure}[t]
    \centering
    \includegraphics[width=1\linewidth]{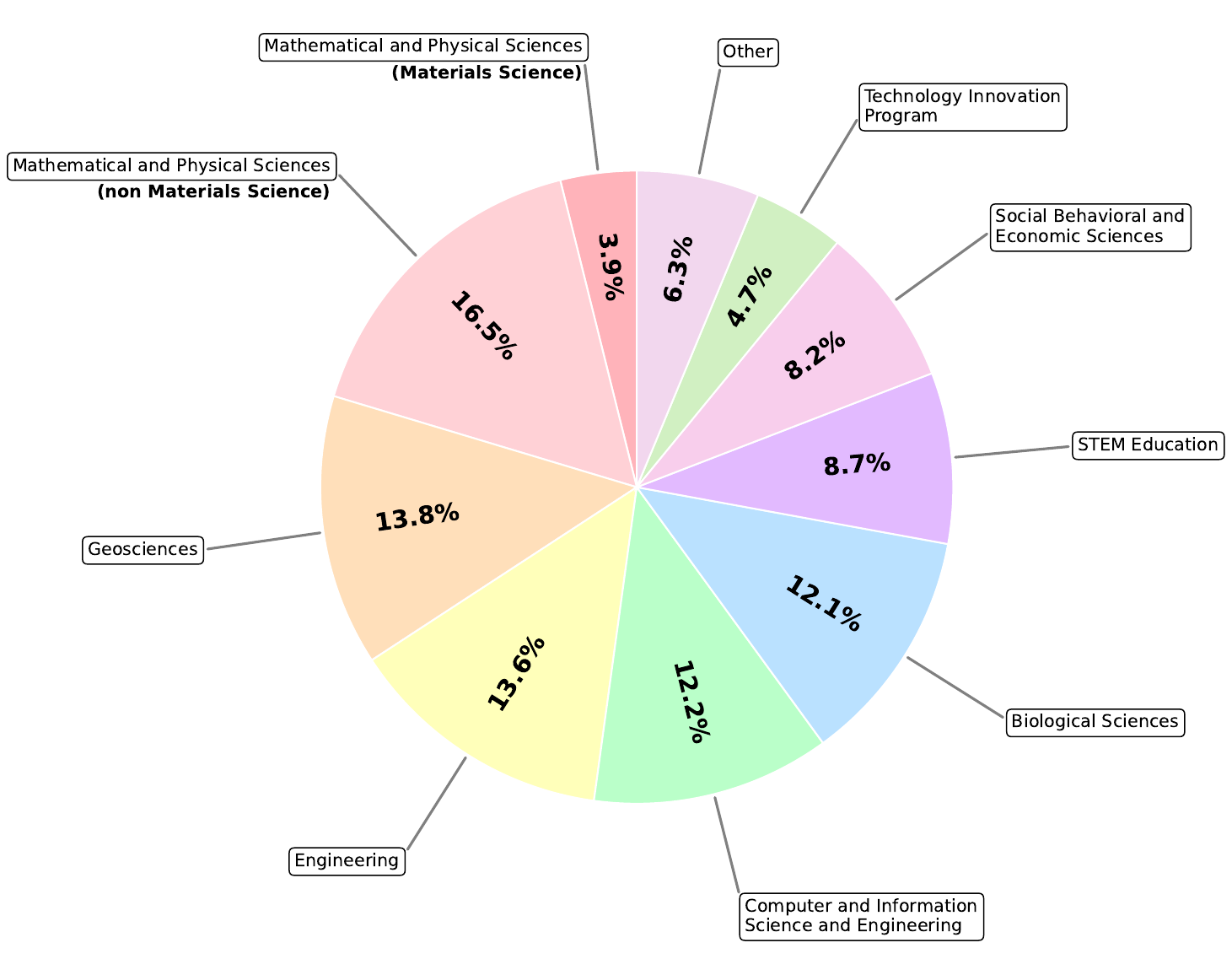}
    \caption{\textbf{(\DatasetName{} contains a large variety of domains.)} Distribution of awards areas as represented by the \NSF~directorates in \DatasetName, illustrating the breadth and comprehensiveness of scientific claims in our dataset. The \DatasetNameMatSci~subset spanning all of materials science awards represents 3.9\% of the entire dataset.}
    \label{fig:award-distribution}
\end{figure}

Table \ref{tab:datasets} situates existing scientific claim datasets with our \DatasetName~datasets, highlighting the significantly larger scale of our contribution (2.8 million claims in \DatasetName, 135,000 claims in\DatasetNameTwentyK~and 114,000 claims in \DatasetNameMatSci), broad topic coverage (all of science and math), and novelty of data source (grant abstracts). See Figure~\ref{fig:award-distribution}.

\paragraph{Meta Science and Social Science} Previous works have examined grants data in social science and meta-science contexts. For example,~\citet{park2024} examine the relationship between interdisciplinary grants and the impact of papers they support and~\citet{xu2022} study the influence of research funding on team structure using grant data. While these are tenuously connected to our work, we list them for the sake of completeness.

\section{Building \DatasetName}

\subsection{Data Collection}
We downloaded the entire NSF Awards database\footnote{\url{https://www.nsf.gov/awardsearch/advancedSearch.jsp}} in XML format, containing more than 0.5 million awards from 1970 through September 2024. After parsing, we obtained 412,155 parseable awards, which we call \DatasetName.

In this paper, we focus on all awards from the Division of Materials Research (DMR), which is responsible for most materials science awards at the NSF.  This subset, called \DatasetNameMatSci, contains 16,031 awards, representing approximately 3.2\% of the entire NSF awards database. We chose materials science as our focus due to its interdisciplinary nature and technological importance. In addition, we build \DatasetNameTwentyK, a different subset of 20K awards spanning 5 NSF directorates --- Mathematical and Physical Sciences (MPS), Geological Sciences (GEO), Engineering (ENG), Computer and Information Science and Engineering (CSE), and Biological Sciences (BIO).

\subsection{Data Processing}

As Figure~\ref{fig:sample-record} illustrates, each record in  \DatasetNameMatSci~typically contains:
\begin{enumerate}[noitemsep,topsep=0pt]
\item Award ID, title, and year.
\item Directorate and division information
\item Technical abstract
\item Non-technical abstract (present in $\sim$81\% of awards)
\item Scientific claims made in the abstracts
\item Investigation proposals in the abstracts
\item Publications resulting from the grant (when available)
\end{enumerate}

The practice of updating awards with resulting publications is relatively recent, primarily occurring from 2014 onwards. For awards where publications are present, we extracted the DOIs and resolved them to obtain titles, abstracts, and publication URLs.

\subsection{Claim and Investigation Proposal Extraction}
\label{sec:claude-claim-extraction}
To extract scientific claims and investigation proposals from the award abstracts, we developed a zero-shot prompting approach using Anthropic's Claude-3.5\footnote{\texttt{Claude-3.5-Sonnet-20240620} accessed between Sep-Oct. 2024, to be specific.} model. Our prompt instructed the model to identify two types of statements:

\begin{enumerate}[noitemsep,topsep=0pt]
\item \textbf{Claims}: Statements that the abstract claims to be true or states as assumptions, either explicitly or implicitly.~\footnote{Our notion of claims follows prior work~\citep{tang-etal-2024-minicheck}.}
\item \textbf{Investigation proposals}: Forward-looking statements that propose specific research activities as part of the award.
\end{enumerate}

We structured the prompt to return a JSON object containing the award ID, technical abstract, non-technical abstract, a list of claims, and a list of investigation proposals. To maintain consistency and quality, we set temperature to zero for all extractions. See Appendix~\ref{appendix:claude-claim-extraction} for the exact prompt and Appendix~\ref{appendix:claim-IP-examples} for sample claims and investigation proposals.

We performed qualitative experiments with several prompt variants and our analysis showed that jointly extracting claims and investigation proposals helped maintain the relevance of extracted claims. When claims were extracted without also extracting investigation proposals, the model often confused forward-looking statements about proposed investigations as factual claims.

\section{Dataset Analysis}
\label{sec:dataset_analysis}
\paragraph{\DatasetName} The full dataset contains 412,155 award abstracts spanning from 1970 to 2024, with 2.8 million scientific claims and corresponding investigation proposals.

\paragraph{\DatasetNameMatSci} This materials science subset, which is the focus of this preprint, contains:
\begin{itemize}[noitemsep,topsep=0pt]
\item 16,042 awards with each with a technical and non-technical abstract
\item 114K extracted scientific claims (average of $7\pm2$ claims per abstract-pair)
\item 145K extracted investigation proposals (average of $9\pm3$ proposals per abstract-pair)
\item 2,953 awards with linked publications (18.4\% of the dataset). Such awards had anywhere between 1 -- 4 publications.
\end{itemize}

\paragraph{\DatasetNameTwentyK} For building models across all NSF~directorates, we take 20,000 sample subset of \DatasetName, by stratifying across 5 directorates.
\begin{itemize}[noitemsep,topsep=0pt]
\item 20,001 awards with each with a technical and non-technical abstract
\item 135K extracted scientific claims (average of $7\pm2$ claims per abstract-pair)
\item 139K extracted investigation proposals (average of $7\pm2$ proposals per abstract-pair)
\end{itemize}

\subsection{Technical vs. Non-Technical Abstracts}
We investigated the differences between technical and non-technical abstracts in our dataset. Using a symmetric BLEU score to measure textual similarity between paired abstracts, we found that only 202 (1.5\%) out of 13,025 technical/non-technical abstract pairs had a similarity score greater than 0.6, suggesting that the non-technical abstracts are not simply copied from the technical abstracts.

Since grant abstracts are previously unexamined in literature, we further investigated the stylistic differences between technical and non-technical abstracts using pre-trained document embedding models. Figure~\ref{fig:tsne-specter-stel} compares content embeddings from SPECTER \cite{cohan-etal-2020-specter} and style embeddings from STEL \cite{patel2025}. Using these embeddings with a linear SVM classifier, we achieved F1 scores of 90.99 (SPECTER), 88.42 (STEL), and 89.99 (concatenated), demonstrating that the abstracts are distinguishable both in content and style.

\subsection{Taxonomies of Claims and Investigation Proposals}
\label{sec:taxonomy}

\begin{figure}[t]
    \centering
    \includegraphics[width=1\linewidth]{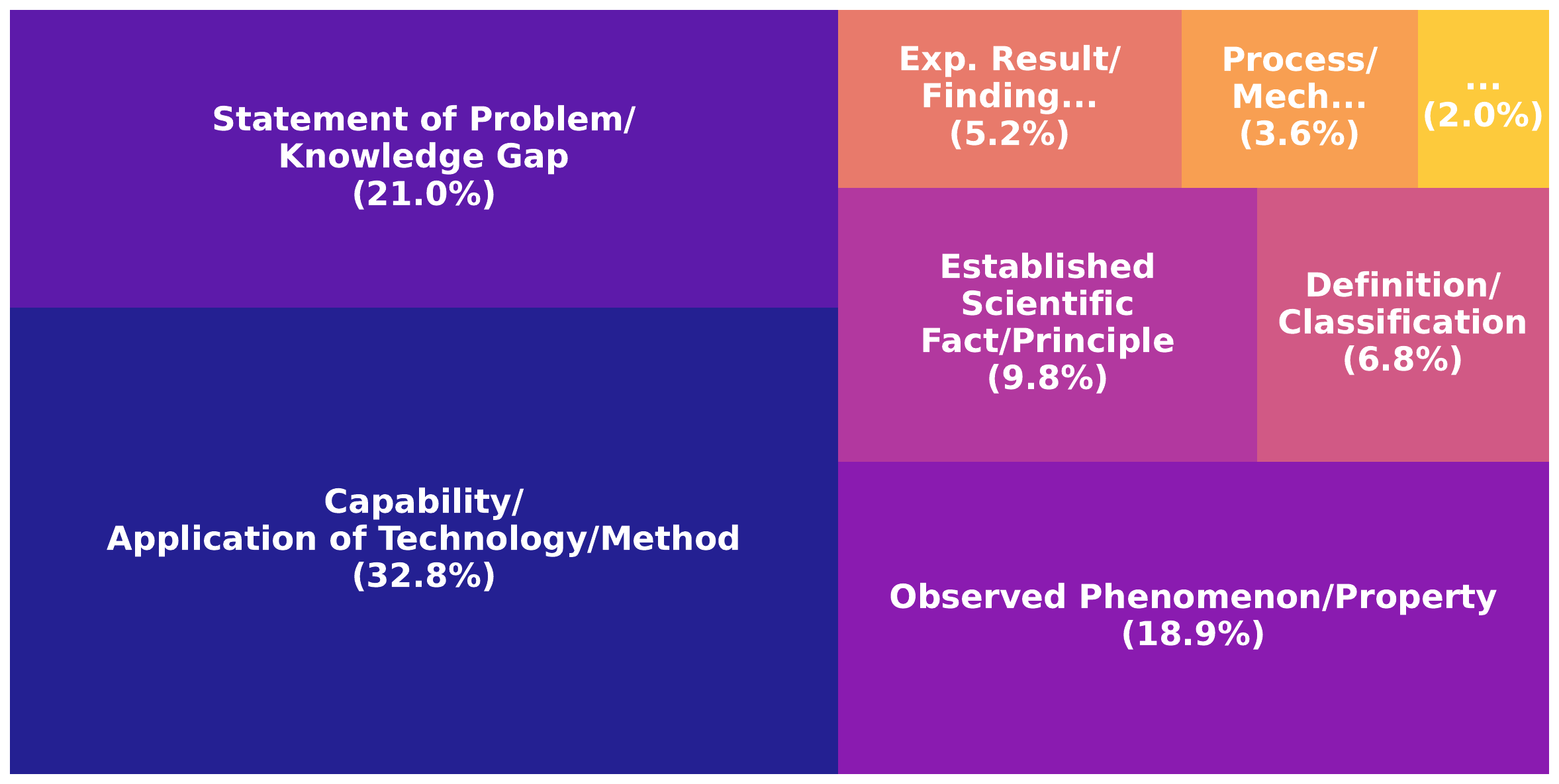}
    \caption{\textbf{(Most scientific claims in the abstracts are about knowledge gap and application methods.)} A treemap of the scientific claim categories in NSF awards. See Table~\ref{tab:claim-cat-examples} for descriptions of these categories.}
    \label{fig:claim-category-treemap}
\end{figure}

\paragraph{Claims.}
To characterize the types of assertions made in NSF award abstracts, we analyzed 810 extracted claims from 120 awards sampled across five NSF directorates (MPS, GEO, ENG, CSE, BIO). We identified eight broad categories, covering well-known facts, observed phenomena, applications of methods or technologies, theoretical predictions, experimental findings, knowledge gaps, definitions/classifications, and process descriptions. Figure~\ref{fig:claim-category-treemap} shows their distribution. The most common types are \textit{Capability/Application of Technology/Method} (32.8\%), \textit{Statement of Problem/Knowledge Gap} (21.0\%), and \textit{Observed Phenomenon/Property} (18.9\%).
Examples for all categories are shown in Table~\ref{tab:claim-cat-examples}.

\begin{figure}[t]
    \centering
    \includegraphics[width=1\linewidth]{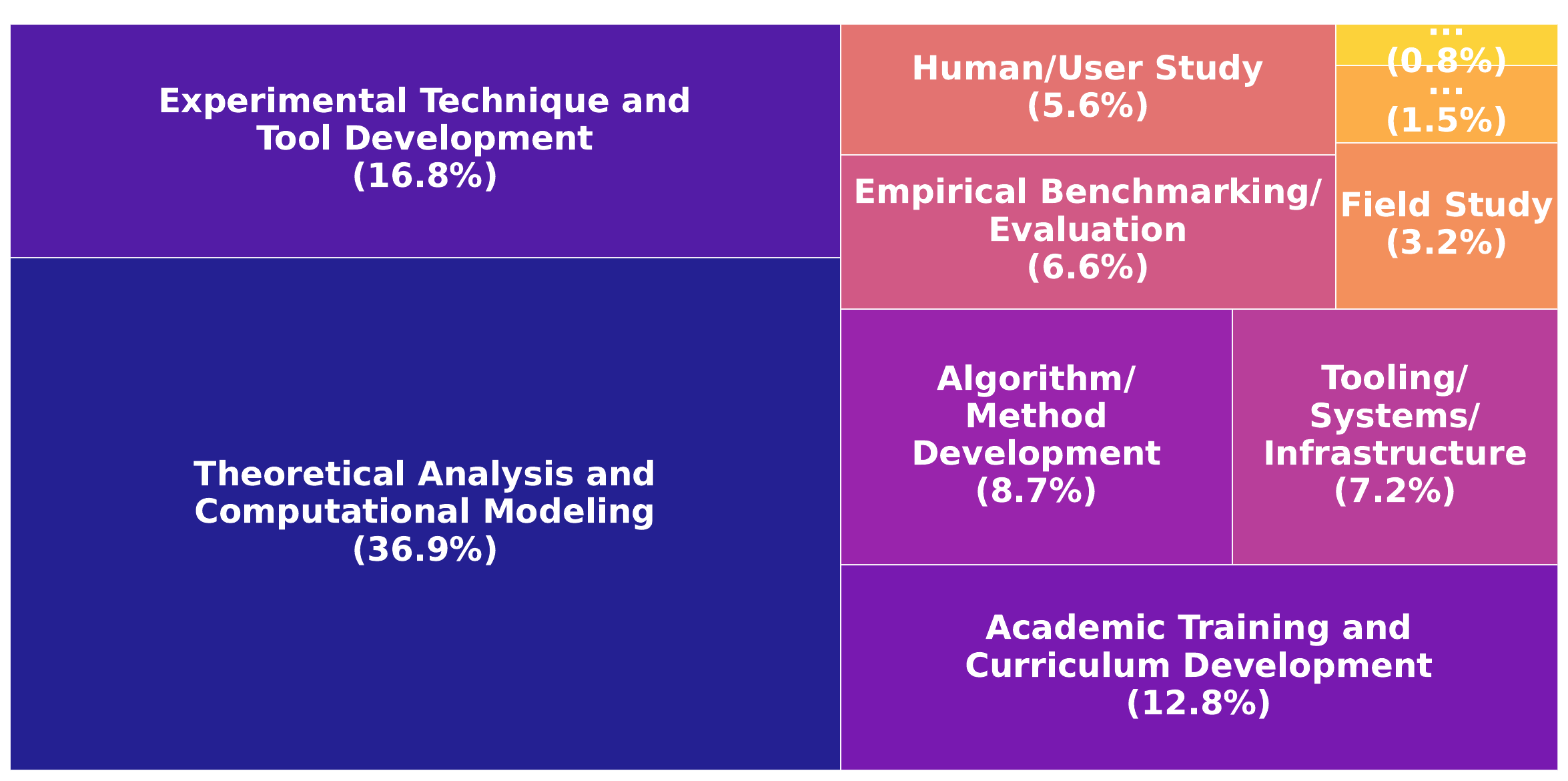}
    \caption{\textbf{(Most investigation proposals in the abstracts are about experimental technique and theoretical analysis.)} A treemap of the investigation proposal categories in NSF awards. See Table~\ref{tab:ip-cat-examples} for descriptions of these categories.}
    \label{fig:ip-category-treemap}
\end{figure}

\paragraph{Investigation Proposals.}
We performed a parallel analysis on 833 investigation proposals from the same award set, identifying eight categories spanning theoretical analysis, experimental technique development, algorithm/method development, academic training, and various empirical study types. Figure~\ref{fig:ip-category-treemap} shows their distribution. The majority fall under \textit{Theoretical Analysis and Computational Modeling} (36.9\%), \textit{Experimental Technique and Tool Development} (16.8\%), and \textit{Academic Training and Curriculum Development} (12.8\%). Examples for all categories are shown in Table~\ref{tab:ip-cat-examples}.

\begin{figure}[t!]
    \centering
    \includegraphics[width=1\linewidth]{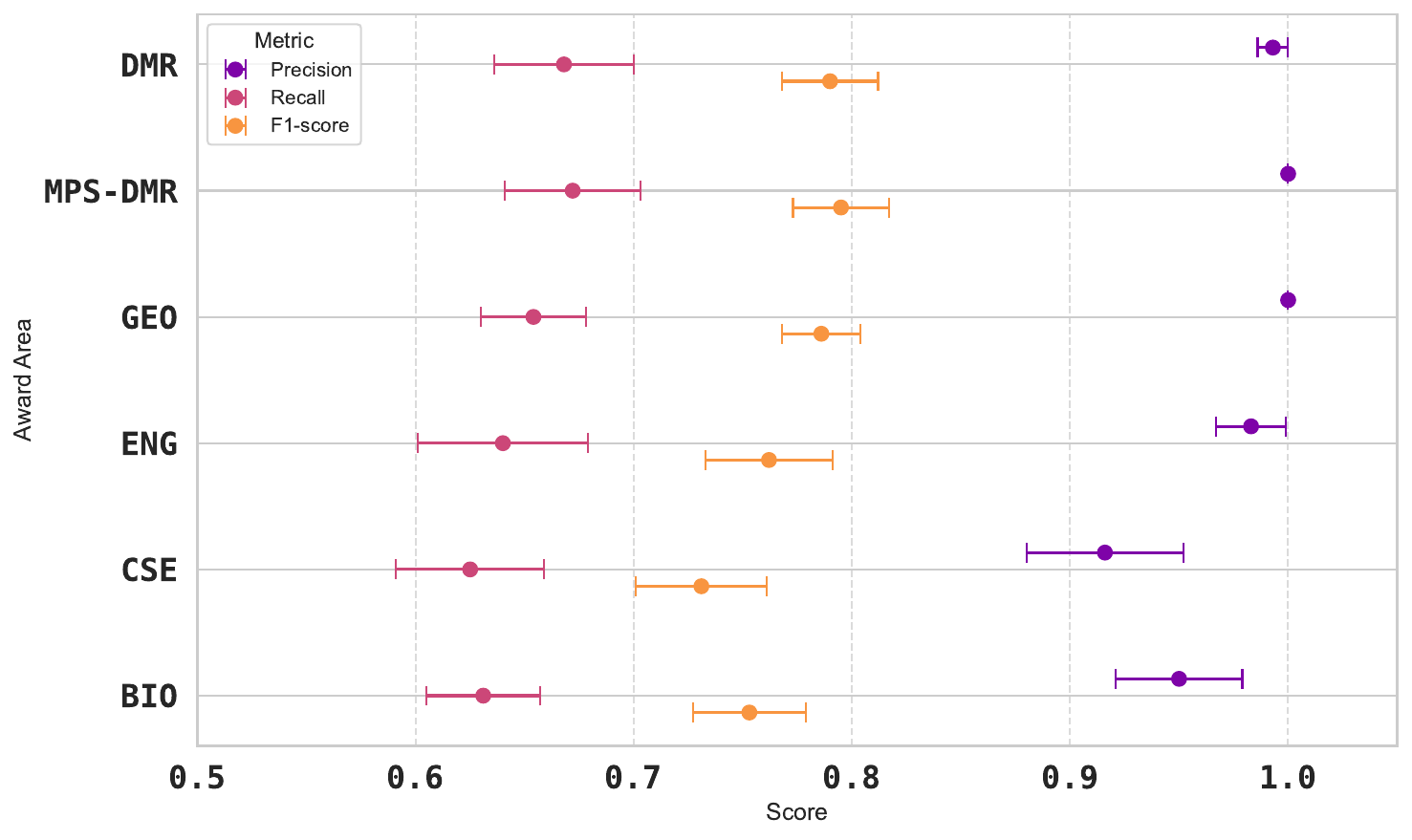}
    \caption{\textbf{(Claim extraction achieves consistently high precision across all areas, while recall is lower, leading to moderate F1-scores.)} A Cleveland dot plot of precision, recall, and F1-score across different NSF Award Areas for claims extracted via Claude (See Section~\ref{sec:claude-claim-extraction}). Error bars denote standard deviation (bootstrap N=1000). See Section~\ref{sec:eval-claude-extraction} for analysis.}
    \label{fig:syndata_eval}
\end{figure}

\begin{figure}[t!]
    \centering
    \includegraphics[width=1\linewidth]{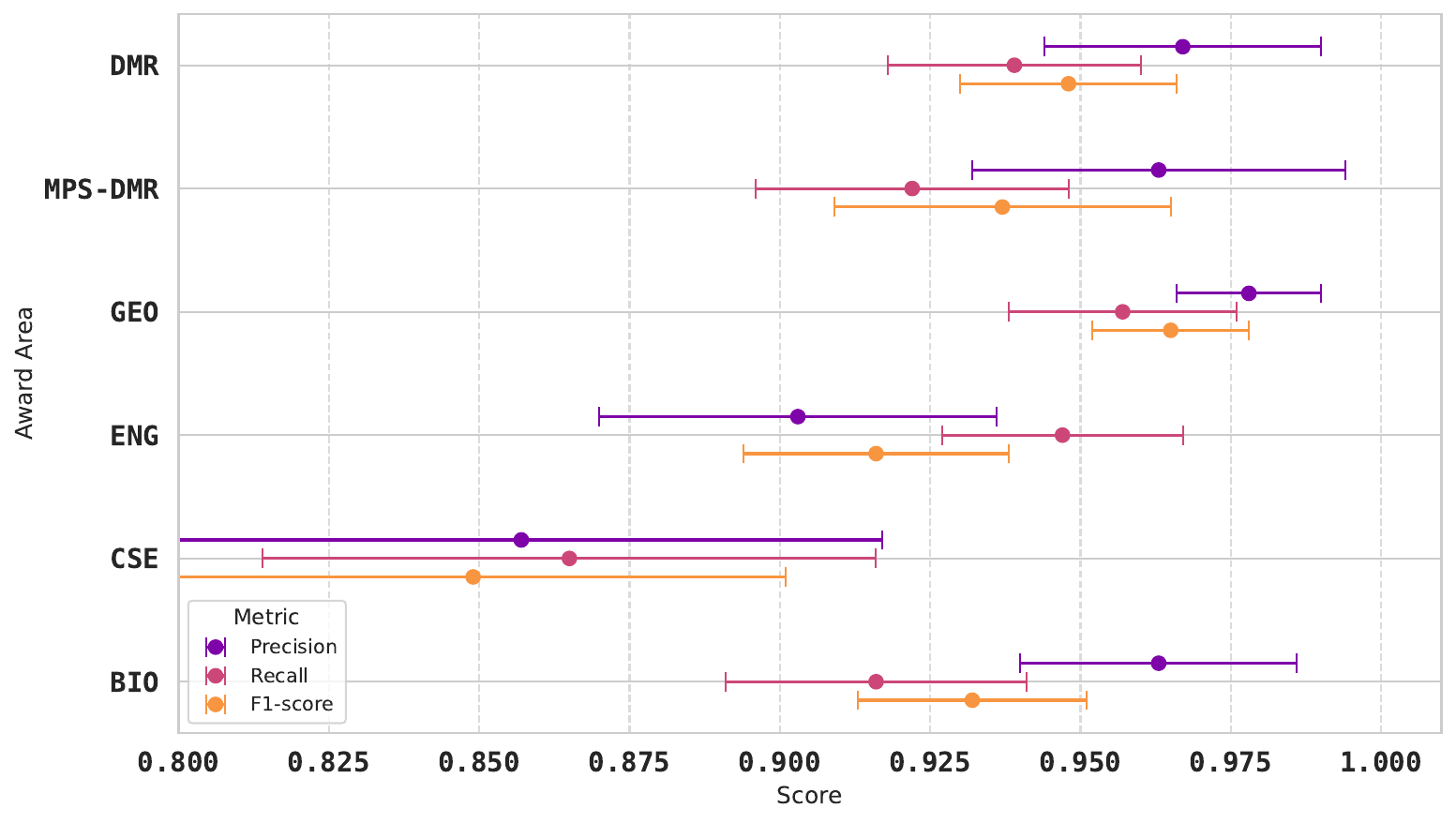}
    \caption{\textbf{(Investigation Proposal extraction achieves consistently high precision across all areas, while recall is lower, leading to moderate F1-scores.)} A Cleveland dot plot of precision, recall, and F1-score across different NSF Award Areas for investigation proposals extracted via Claude (See Section~\ref{sec:claude-claim-extraction}). Error bars denote standard deviation (bootstrap N=1000). See Section~\ref{sec:eval-claude-extraction} for analysis.}
    \label{fig:syndata_eval_ip}
\end{figure}


\subsection{Evaluating Extracted Claims and Investigation Proposals}
\label{sec:eval-claude-extraction}

We evaluate the quality of the extracted claims and investigation proposals (Section~\ref{sec:claude-claim-extraction}) by manually annotating 120 sampled awards (Section~\ref{sec:taxonomy}) and computing precision, recall, and F1. For each of the six NSF areas---Materials Science (DMR), Mathematical and Physical Sciences excluding Materials Science (MPS-DMR), Geological Sciences (GEO), Engineering (ENG), Computer and Information Science and Engineering (CSE), and Biological Sciences (BIO)---we randomly sampled 20 items per area. Using \texttt{GPT-4o}~\citep{openai2024gpt4ocard}, we identified additional true elements $G'$ missed by the extracted set (with $FN = |G'|$) and categorized previously extracted elements as correct ($TP$) or incorrect ($FP$). Annotators (PhD students) manually verified \texttt{GPT-4o}’s outputs on 20 abstracts and confirmed near-perfect verification accuracy. Precision, recall, and F1 were then computed using $FN$, $TP$, and $FP$.

Figures~\ref{fig:syndata_eval} and~\ref{fig:syndata_eval_ip} summarize performance across the six areas for claims and investigation proposals, respectively. For claims, extraction achieves consistently high precision but lower recall, leading to moderate F1-scores. For investigation proposals, precision, recall, and F1 are more balanced across areas, indicating more comprehensive coverage. Overall, the extracted data is of high quality, though improving recall for claims remains an important direction.



\section{Tasks, Metrics, and Experiments}
\label{sec:tasks-metrics-experiments}

Previously, Section~\ref{sec:claude-claim-extraction} describes the data extraction process using a large model, and Section~\ref{sec:dataset_analysis} evaluates the quality of the resulting synthetic data. Here, we demonstrate its utility by evaluating the performance of smaller models fine-tuned on it across three NLP tasks:


\begin{enumerate}[noitemsep, wide, labelwidth=!, labelindent=0pt]
\item The \textbf{Non-technical Abstract Generation} task translates dense, technical grant abstracts into accessible language for broader science communication. Motivated by capturing the core scientific essence while navigating stylistic and content differences between technical and lay summaries, this task uses the dataset's paired examples (common in NSF awards) to train models for this nuanced transformation.
\item The \textbf{Abstract to Scientific Claims Extraction} task automates identifying verifiable assertions—the core of scientific discourse—from grant abstracts, which capture these claims at an early, pre-publication stage. Significant performance gains post-fine-tuning highlight the dataset's effectiveness in teaching models to pinpoint these crucial statements.
\item The \textbf{Abstract to Investigation Proposals Extraction} task distinguishes aspirational research intentions from established claims, offering a novel analysis of scientific texts. This provides a clearer view of the planned research trajectory by identifying intended activities. It complements claim extraction by presenting a fuller picture of proposed work, from assertions to investigative pathways, again showing significant fine-tuning efficacy due to the dataset's focused nature.
\end{enumerate}

To explore the three tasks, we finetuned two 7B parameter language models:
\begin{itemize}[noitemsep,topsep=0pt]
\item \texttt{Mistral-7B-instruct-v0.3}~\cite{jiang2023mistral7b}
\item \texttt{Qwen2.5-7B-Instruct}~\cite{qwen2}
\end{itemize}
\noindent 

\subsection{Data Preparation}

Starting with 16,042 processed entries in \DatasetNameMatSci, we removed near-duplicates in technical and non-technical abstracts using trigram Jaccard similarity (threshold > 0.9), resulting in 11,569 data points. We further filtered cases where character-level 10-gram similarity between an entry's technical and non-technical abstracts exceeded 0.6, yielding 11,141 final data points. We split this dataset into train/validation/test sets with 8,641/500/2,000 examples, respectively.

\subsection{Finetuning Details}
For fine-tuning, we used LoRA~\cite{hu2021lora} with rank=128, lora\_alpha=64 and a learning rate of 1e-5 scheduled linearly. We updated the query, key, value, and output projection layers, as well as MLP gate, up, and down projections. We ran the finetuning on an A100 GPU for 3 epochs, 100 warmup steps, and a batch size of 2 with 4 accumulated steps.
Each epoch takes around one hour.

\subsection{Evaluation Metrics}

For Task 1 -- abstract generation -- we employed a comprehensive evaluation framework using both BERTScore~\cite{zhang2020bertscore} and ROUGE~\cite{lin-2004-rouge} metrics to assess the quality of generated non-technical abstracts. This combination enables us to capture both lexical overlap and structural similarity through the ROUGE variants, while BERTScore provides insights into semantic alignment between the generated texts and reference abstracts. Incorporating such multi-viewed metrics\footnote{For BERTScore we report precision, recall and F1, and for ROUGE we report ROUGE-1, ROUGE-2, ROUGE-L, and ROUGE-L-sum.} ensures that the evaluation reflects not only the presence of key words and phrases but also the underlying meaning and narrative coherence of the abstracts.

For Task 2 -- claim extraction -- we developed a novel evaluation approach using LLM-based comparisons. Previous methods for claim evaluations focused on comparing a single claim against a single document. See ~\citet{tang-etal-2024-minicheck}, for example. However, our setting required evaluating a set of extracted claims against a gold set of claims.

Towards that end, we defined a boolean function $\mathbf{\Phi}_{\textrm{claim}}$ using \texttt{GPT-4o-mini}~\citep{openai2024gpt4omini} with zero-shot prompting to determine whether a generated claim is supported by a gold standard claim. See Appendix~\ref{appendix:phi_claim} for prompt details\footnote{We tried several slight edits of the prompts and found them to be robust to such changes.}. Using this function, we calculated precision and recall as follows:
\begin{equation*}
\left.\begin{aligned}
\text{Precision}&=&\frac{1}{|S|} \sum_{c \in S} \max_{g \in G} \mathbf{\Phi}_{\textrm{claim}}(c, g)\\
\text{Recall}&=&\frac{1}{|G|} \sum_{g \in G} \max_{c \in S} \mathbf{\Phi}_{\textrm{claim}}(g, c)
\end{aligned}\right.
\end{equation*}
\noindent where $S$ is the set of claims generated from the finetuned model, after removal of any repeats/near-repeats
\footnote{We determine repeats and near-repeats in the generation by thresholding cosine similarity calculated over a TF-IDF representation of the generated claims.},
 and $G$ is the gold standard set. We note that this is a variant of precision/recall metrics defined for image captioning in~\cite{deitke2024molmo}, however unlike ~\citeauthor{deitke2024molmo}, we explicitly use $\mathbf{\Phi}_{\textrm{claim}}$ in computing both precision and recall. This is necessary as we need to accurately penalize any spurious claims generated by the finetuned model. Works by ~\cite{gu2025surveyllmasajudge, liu2023g} are relevant here.

We carefully validated our LLM on a subset of 120 awards using human annotators assisted by \texttt{GPT-4o-mini}. We restricted the role of GPT-4o-mini to only pairwise sentence comparison, a task which prior work has shown as easy for large foundation models. We found a near-perfect correlation between human judgments and \texttt{GPT-4o-mini}'s judgements for this pairwise comparison~\footnote{We use \texttt{GPT-4o-mini} here because this is a simple task and we found \texttt{GPT-4o-mini} sufficient.}. Based on this validation, we applied LLM-as-judge evaluation to the full dataset, a scale that would otherwise have been infeasible to annotate manually. All P/R/F1 values were computed deterministically using the pairwise outputs.

Analogously, for Task 3 -- extraction of investigation proposals -- we define precision and recall similarly but use a different pairwise boolean judge function $\mathbf{\Phi}_{\textrm{IP}}$ \textit{mutatis mutandis}. See Appendix~\ref{appendix:phi_IP} for prompt details.

\section{Results}

\subsection{Non-technical Abstract Generation}
Table \ref{tab:eval_tech2nontech} shows the results for Task 1. Both Mistral and Qwen models demonstrated strong performance, with fine-tuning providing modest improvements. The Mistral model outperformed Qwen on almost all metrics, achieving a BERTScore-F1 of 0.8561 after fine-tuning (+0.36\% relative improvement). ROUGE scores were generally low (0.01-0.22), reflecting the stylistic differences between technical and non-technical abstracts.

\begin{table}[t]
\centering
\small
\resizebox{\linewidth}{!}{%
\begin{tabular}{lll}
\toprule
Metric & Mistral & Qwen\\
\midrule
BERTScore-P & 0.8563 \textcolor{darkgreen}{(+0.38\% $\uparrow$)} & 0.8459 \textcolor{darkgreen}{(+0.98\% $\uparrow$)} \\
BERTScore-R    & 0.8555 \textcolor{darkgreen}{(+0.30\% $\uparrow$)} & 0.8597 \textcolor{darkgreen}{(+1.61\% $\uparrow$)} \\
BERTScore-F1        & \textbf{0.8561} \textcolor{darkgreen}{(+0.36\% $\uparrow$)} & 0.8437 \textcolor{darkgreen}{(+0.75\% $\uparrow$)} \\
ROUGE1              & 0.2000 \textcolor{darkgreen}{(+2.58\% $\uparrow$)} & 0.1978 \textcolor{darkgreen}{(+1.98\% $\uparrow$)} \\
ROUGE2              & 0.0198 \textcolor{darkgreen}{(+4.76\% $\uparrow$)} & 0.0210 \textcolor{darkgreen}{(+3.89\% $\uparrow$)} \\
ROUGE-L             & 0.1273 \textcolor{darkgreen}{(+2.96\% $\uparrow$)} & 0.1466 \textcolor{darkgreen}{(+0.65\% $\uparrow$)} \\
ROUGE-L-sum         & 0.2166 \textcolor{darkgreen}{(+2.45\% $\uparrow$)} & 0.2078 \textcolor{darkgreen}{(+1.66\% $\uparrow$)} \\
\bottomrule
\end{tabular}
}
\caption{\textbf{(Finetuned models have modest improvements on technical abstract to non-technical abstract translation, indicating excellent out-of-the-box performance for this task.)} Finetuning performance for \texttt{Mistral-7B-instruct-v0.3}
 and \texttt{Qwen2.5-7B-Instruct}
 models for Technical abstract to Non-technical abstract translation (Task 1), with relative improvements over the corresponding unfinetuned model indicated in \textcolor{darkgreen}{green}. Error bars for all metrics at 95\% confidence intervals range between 0.0000--0.0025. Mistral model outperforms Qwen on almost all metrics for this task regardless of finetuning. 
 }
\label{tab:eval_tech2nontech}
\end{table}

\subsection{Scientific Claim Extraction}

\begin{table}[t]
\centering
\small
\begin{tabular}{lll}
\toprule
Metric & Mistral & Qwen \\
\midrule
Precision
 & 0.7450 \textcolor{darkgreen}{\,(+116.7\% $\uparrow$)} 
 & 0.6839 \textcolor{darkgreen}{\,(+107.1\% $\uparrow$)} \\
Recall 
 & 0.7098 \textcolor{darkgreen}{\,(+59.5\% $\uparrow$)} 
 & 0.6611 \textcolor{darkgreen}{\,(+7.8\% $\uparrow$)} \\
F1 
 & 0.7097 \textcolor{darkgreen}{\,(+101.8\% $\uparrow$)} 
 & 0.6541 \textcolor{darkgreen}{\,(+63.3\% $\uparrow$)} \\
\bottomrule
\end{tabular}
\caption{\textbf{(Finetuning leads to large improvements in claim extraction from abstracts.)} Finetuning performance for \texttt{Mistral-7B-instruct-v0.3}
 and \texttt{Qwen2.5-7B-Instruct}
 models for Claim Extraction from abstracts (Task 2), with relative improvements over the corresponding unfinetuned model indicated in \textcolor{darkgreen}{green}. Error bars for all metrics at 95\% confidence intervals range between 0.0038--0.0055. Mistral model outperforms Qwen on almost all metrics for this task regardless of finetuning. We note the large positive percent changes, sometimes improvements as large as 2x, indicate finetuning is indispensable for claim extraction. Mistral model outperforms Qwen on almost all metrics for this task.}
\label{tab:eval_technontech2claims}
\end{table}

For Task 2 (claim extraction), fine-tuning yielded substantial improvements. As shown in Table \ref{tab:eval_technontech2claims}, the fine-tuned Mistral model achieved a precision of 0.7450 (+116.7\% relative improvement), recall of 0.7098 (+59.5\%), and F1 of 0.7097 (+101.8\%). The Mistral model consistently outperformed Qwen, though both showed significant benefits from fine-tuning.

\subsection{Investigation Proposal Extraction}
Similarly, Task 3 (proposal extraction) showed dramatic improvements with fine-tuning. As shown in Table \ref{tab:eval_technontech2ip}, the Mistral model achieved a precision of 0.7351 (+18.24\%), recall of 0.7539 (+127.24\%), and F1 of 0.7261 (+90.97\%) after fine-tuning. The relative improvements were even larger for the Qwen model, though Mistral still performed better overall.

Since Mistral models seemed to have an edge over the Qwen2.5 models for these tasks, we also trained a Mistral only version of on the \DatasetNameTwentyK~subset which spans all NSF directorates. The results can be found in Appendix~\ref{appendix:20k-eval}.

\section{Error Analysis}
\label{sec:error-analysis}
We conduct error analyses on both claim extraction and investigation proposal extraction to understand common failure modes of fine-tuned models.

\paragraph{Claims.}
\label{sec:error-analysis-claims}
Using 120 awards from the test sets of \DatasetNameMatSci{} and \DatasetNameTwentyK{}, we examined 802 claims generated by a fine-tuned Mistral-7B model and found an error rate of 2.6\%. We categorized the errors into five types: (1) \textbf{Overconfidence} — misrepresenting hedged statements as factual assertions; (2) \textbf{Mixing Information} — combining content from multiple sentences incorrectly; (3) \textbf{Overgeneralization} — extending claims beyond what is stated; (4) \textbf{Information Omission} — dropping key qualifiers and altering meaning; and (5) \textbf{Administrative Hallucinations} — inserting funding or institutional information not present. Overconfidence and overgeneralization were the most common. Claude-extracted claims had a slightly lower error rate (2.1\%), mostly administrative hallucinations.

\begin{table}[t]
\centering
{\small
\begin{tabular}{lll}
\toprule
Metric & Mistral & Qwen\\
\midrule
Precision
 & 0.7351 \textcolor{darkgreen}{(+18.24\% $\uparrow$)} 
 & 0.7245 \textcolor{darkgreen}{(+70.07\% $\uparrow$)} \\
Recall 
 & 0.7539 \textcolor{darkgreen}{(+127.24\% $\uparrow$)} 
 & 0.6865 \textcolor{darkgreen}{(+81.57\% $\uparrow$)} \\
F1
 & 0.7261 \textcolor{darkgreen}{(+90.97\% $\uparrow$)} 
 & 0.6827 \textcolor{darkgreen}{(+112.60\% $\uparrow$)} \\
\bottomrule
\end{tabular}
\caption{\textbf{(Finetuning leads to large improvements in investigation proposal extraction from abstracts.)} Finetuning performance for \texttt{Mistral-7B-instruct-v0.3}
 and \texttt{Qwen2.5-7B-Instruct}
 models for extraction of Investigation Proposals from award abstracts (Task 3), with relative improvements over the corresponding unfinetuned model indicated in \textcolor{darkgreen}{green}. Error bars for all metrics at 95\% confidence intervals range between 0.0036--0.0073. Mistral model outperforms Qwen on almost all metrics for this task regardless of finetuning. We note the large positive percent changes, sometimes improvements as large as 2x, indicate finetuning is indispensable for this task. Mistral model outperforms Qwen on almost all metrics for this task.}
\label{tab:eval_technontech2ip}
}
\end{table}

\paragraph{Investigation Proposals.}
\label{sec:error-analysis-ip}
A parallel analysis on 833 investigation proposals yielded an error rate of 2.4\%. We identified four error types: (1) \textbf{No Investigation Proposals} — generating proposals when none exist in the abstract; (2) \textbf{Content Mismatch} — introducing or omitting key elements; (3) \textbf{Overspecification} — adding unsupported details; and (4) \textbf{Existing Work} — describing prior work rather than forward-looking plans.

Examples per error type are in Appendix~\ref{app:error-examples}. Mitigation strategies across both tasks include uncertainty calibration, 
and stricter alignment between extractions and source text.
We manually check 20 examples and found most ``correct'' claims are indeed correct, while over half of the ``errors'' are not actual errors, suggesting even higher true accuracy.

\section{Discussion and Conclusion}
\label{sec:conclusion}
We introduced \DatasetName, a large dataset of 2.8 million scientific claims and proposals from 400,000 NSF grant abstracts across all science and mathematics disciplines. Focused subsets include \DatasetNameMatSci~(114,000 materials science claims) and \DatasetNameTwentyK~(135,000 claims from five directorates).
Experiments demonstrate that fine-tuning language models on \DatasetName{} significantly improves scientific claim and proposal extraction, with relative performance gains often exceeding 100\%. Non-technical abstract generation saw modest improvements due to strong baselines.
Stylistic differences between technical and non-technical abstracts offer potential for science communication. Our claim taxonomy identifies prevalent assertion types like capability/application and problem/knowledge gap statements.
\DatasetName's unique advantages include its vast scale, high quality from NSF expert review, comprehensive coverage of scientific domains, a temporal span from 1970-2024 enabling longitudinal studies, and, for recent grants, links to resulting publications.
\DatasetName~opens new research avenues in large-scale claim verification, scientific discovery tracking, and meta-scientific analysis, a key resource for understanding scientific assertions at their origin.

\section*{Limitations}
\paragraph{Source Material Scope.} The dataset, derived from NSF award abstracts, offers insights into early-stage scientific claims from a rigorously reviewed, cross-disciplinary source. However, it currently excludes claims from unfunded proposals or international contexts. Future work may expand to other agencies and sources.

\paragraph{Bias and Coverage Considerations.} 
While the dataset currently excludes unfunded and international proposals, the National Science Foundation accounts for approximately 25\% of U.S. federally supported basic research, providing substantial coverage across scientific disciplines. We also note an availability bias: (1) unfunded proposals are not publicly accessible, aside from a handful of exemplars shared online, and (2) international proposals are rare and geographically dispersed. Given their importance, systematically incorporating international proposals represents an important direction for future work.


\paragraph{Extraction Methodology.} Our approach utilizes zero-shot prompting with large language models, refined by prompt engineering and selective human validation. While manual evaluation shows consistently high precision across all directorates, our zero-shot extraction pipeline exhibits lower recall. At this bootstrapping stage, this was a deliberate design choice -- we prioritized high precision to ensure the foundational reliability of the extracted statements and to prevent the proliferation of spurious claims. As demonstrated in Table~\ref{tab:eval_technontech2claims}, fine-tuning smaller models on this dataset significantly improves extraction, roughly doubling the F1 score for both claim and proposal tasks by significantly boosting recall alongside precision. Furthermore, the massive scale of \DatasetName~enables data-intensive strategies to close the recall gap in future work. For instance, the dataset's massive size and cross-disciplinary diversity provide the necessary training signals for multi-pass extraction protocols, allowing models to iteratively capture secondary claims. It also serves as a robust foundation for fine-tuning diverse open-source models for agreement-based ensembling. Finally, the vast candidate pool allows for targeted active annotation, enabling researchers to isolate and manually label only the most complex, low-confidence edge cases to systematically improve recall.

\paragraph{Evaluation Design.} We introduced LLM-based metrics for evaluating claims and investigation proposals, offering a nuanced assessment beyond lexical overlap. These metrics correlate well with human judgment in samples, but broader validation across more scientific domains is needed to confirm their robustness. The public dataset and code aim to facilitate such community efforts.

\paragraph{Temporal and Linked Data Coverage.} Spanning over five decades and including recent linked publication metadata, the dataset's systematic outcome tracking is limited for older awards. This restricts longitudinal analysis of claim evolution from proposal to publication. Broader, consistent outcome reporting could enrich \DatasetName~for deeper research trajectory studies.

\paragraph{Generalizability.} While designed and validated for \NSF~abstracts, whose structure may differ from other scientific communications, the general framework is adaptable. It could be extended to related corpora like other funding agencies, patent abstracts, or scientific news, creating opportunities for future research.

\paragraph{Baselines.}
We report results using two competitive baseline models --- Mistral-7B-v0.3 and Qwen2.5-7B --- and observe consistent trends across both. We do not include additional baselines in this work; a more extensive comparison with other models is left for future work. All datasets and models are publicly released to facilitate such comparisons.

\section*{Acknowledgments}
The authors would like to acknowledge NSF award CCF \texttt{2442421}, the AI2050 program at Schmidt Sciences (Grant \texttt{G-25-67983}), the Defense Advanced Research Projects Agency (DARPA) SciFy program (Agreement No. \texttt{HR00112520300}) for funding this research, and the Office of the Director of National Intelligence (ODNI), Intelligence Advanced Research Projects Activity (IARPA), via 56000026C0019. We also thank the National Science Foundation (NSF) for making award data publicly available, enabling this research. Any views, opinions, findings and conclusions or recommendations expressed in this material are those of the author(s) and do not necessarily reflect the official policy, position, or views, either expressed or implied, of the National Science Foundation, DARPA, the Department of Defense, ODNI, IARPA, or the U.S. Government.
The U.S. Government is authorized to reproduce and distribute reprints for governmental purposes notwithstanding any copyright annotation therein.

\bibliography{main}

\section*{Appendix}

\appendix
\renewcommand{\thetable}{A\arabic{table}}
\renewcommand{\thefigure}{A\arabic{figure}}

\section{Reproducibility Statement}
\label{app:reproduce}
To foster research on large-scale claim extraction, we are releasing our datasets, training code, and trained models:
\begin{itemize}[noitemsep,topsep=0pt]
\item \DatasetNameMatSci: Materials Science subset with extracted claims, investigation proposals, and resolved publication information.
\item \DatasetName: Similar in content to~\DatasetNameMatSci, but a larger superset spanning all of NSF awards database. The key difference is the claims and investigation proposals are extracted from our finetuned models instead of frontier LLMs.
\item Our code is available at \url{https://github.com/darpa-scify/NSFSciFy}.
\item Our best finetuned model checkpoints for extraction of claims and investigation proposals at 
\url{https://huggingface.co/darpa-scify/nsf-scify-matsci-claims}.
\item License: We will release our data and model under apache-2.0.
\item We used all existing artifacts in accordance with their intended research purposes, and we specify that \DatasetName{} is released solely for research and commercial use under compatible access conditions.
\end{itemize}

\section{Complete Prompt for Extracting Claims and Investigation Proposals}
\label{appendix:claude-claim-extraction}
\begin{mdframed}[backgroundcolor=blue!20]
\noindent You are an expert materials science researcher. Given an input JSON description of an NSF material science award abstract, parse out the technical and nontechnical abstracts, and identify the claims and research/investigation proposals the abstract makes. Be thorough. Answer in the following JSON format:
\begin{lstlisting}
{
  "award_id": "", // copied from input
  "technical_abstract": ""  // technical abstract if present, otherwise contents of the abstract field in the input
  "non_technical_abstract": /non-technical abstract if present, otherwise empty
  "claims": [ // list of strings
  ],
  "investigation_proposals":  [ // list of strings
  ],
}    
\end{lstlisting}

\noindent claims are statements that the abstract claims to be true or states as an assumption explicitly or implicitly. \\
\noindent investigation\_proposals are forward-looking statements that the abstract proposals to investigate as a part of this award. \\
\noindent Ensure that the output is in JSON format and that the JSON is valid.
\end{mdframed}
We manually tested the prompt with a few award abstracts to make sure it was optimal for this task.

\section{Prompt for Task 2 evaluation function $\mathbf{\Phi}_{\textrm{claim}}$}
\label{appendix:phi_claim}
\begin{mdframed}[backgroundcolor=blue!20]
\noindent Check two scientific claims c1 and c2, if c1 is supported by c2. If c2 includes all the evidences for c1, but also includes additional content, then it should still be supported (YES). If not all information of c1 is included in c2, or if c2 contains information that conflicts with information in c1, then it should be unsupported (NO). Answer only as a YES or NO.\\
\noindent c1: \{c1\}\\
\noindent c2: \{c2\}
\end{mdframed}

\section{Prompt for Task 3 evaluation function $\mathbf{\Phi}_{\textrm{IP}}$}
\label{appendix:phi_IP}
\begin{mdframed}[backgroundcolor=blue!20]
\noindent Check two investigation proposals c1 and c2, if c1 is supported by c2. If c2 includes all the investigations proposed by c1, but also includes additional proposals, then it should still be supported (YES). If not all proposed investigations by c1 is included in c2, or if c2 contains investigation actions that conflict with investigation actions in c1, then it should be unsupported (NO). Answer only as a YES or NO.\\
\noindent c1: \{c1\}\\
\noindent c2: \{c2\}
\end{mdframed}

\section{Stylistic Differences between Technical and Nontechinal Abstracts}
\begin{figure*}[b!]
    \centering
    \includegraphics[width=1\linewidth]{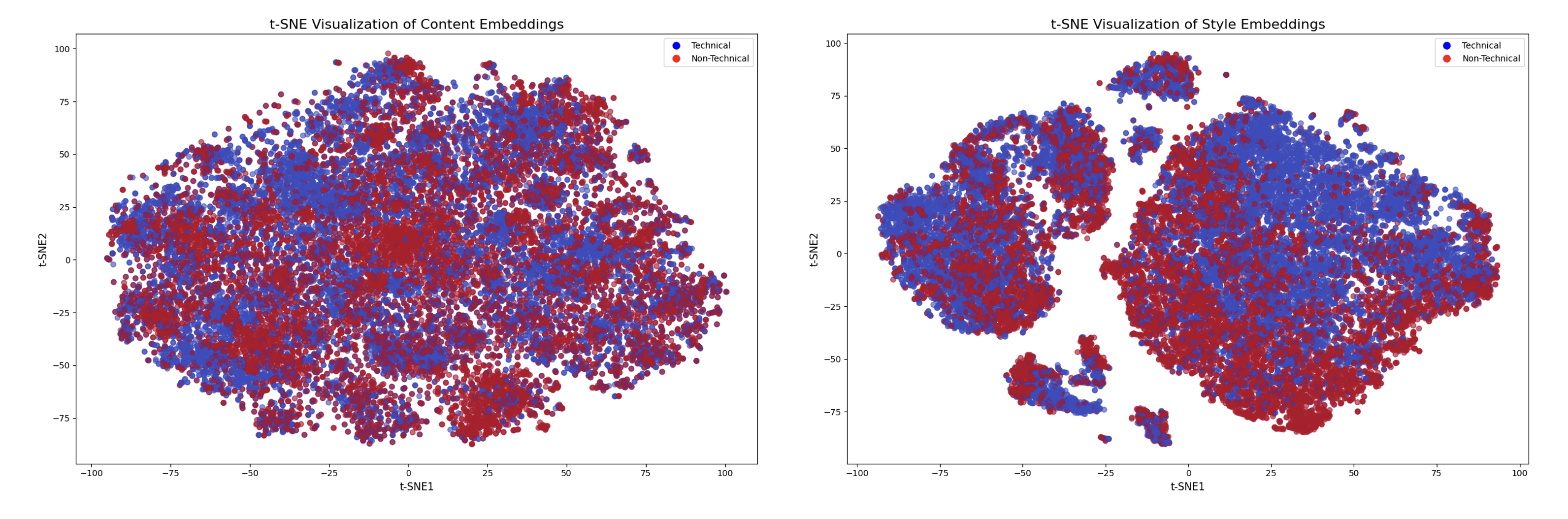}
    \caption{The t-SNE plot of comparing content embeddings from SPECTER \cite{cohan-etal-2020-specter} and style embeddings from STEL \cite{patel2025} for technical and non-technical abstracts in \DatasetNameMatSci. The somewhat clear separation between technical and non-technical abstracts when using style embeddings indicate marked stylistic differences between the two kinds abstracts.}
    \label{fig:tsne-specter-stel}
\end{figure*}

Figure~\ref{fig:tsne-specter-stel} shows stylistic differences between technical and nontechnical abstracts. 

\section{Evaluation results for \DatasetNameTwentyK}
\label{appendix:20k-eval}
Tables \ref{tab:tech2nontech}, \ref{tab:text2claims}, and \ref{tab:text2ip} summarize the results for the three generation tasks defined in Section~\ref{sec:tasks-metrics-experiments} on \DatasetNameTwentyK.

\begin{table}[H]
\centering
{\scriptsize
\begin{tabular}{lrr}
\toprule
Model & Base & Finetuned \\
Metric & & \\
\midrule
BERTScore-F1         & 0.8514 $\pm$ 0.0003 & 0.8500 $\pm$ 0.0006 \\
BERTScore-Precision  & 0.8515 $\pm$ 0.0003 & 0.8513 $\pm$ 0.0007 \\
BERTScore-Recall     & 0.8516 $\pm$ 0.0003 & 0.8496 $\pm$ 0.0005 \\
ROUGE-rouge1         & 0.3351 $\pm$ 0.0013 & 0.3141 $\pm$ 0.0023 \\
ROUGE-rouge2         & 0.0705 $\pm$ 0.0008 & 0.0936 $\pm$ 0.0016 \\
ROUGE-rougeL         & 0.1773 $\pm$ 0.0008 & 0.1967 $\pm$ 0.0016 \\
ROUGE-rougeLsum      & 0.1982 $\pm$ 0.0010 & 0.1998 $\pm$ 0.0016 \\
\bottomrule
\end{tabular}
}
\caption{Technical to Non-Technical Abstract Task: Mistral-7B}
\label{tab:tech2nontech}
\end{table}

\begin{table}[H]
\centering
{\small
\begin{tabular}{lrr}
\toprule
Model & Base & Finetuned \\
\midrule
Precision & 0.4146 $\pm$ 0.0025 & 0.7526 $\pm$ 0.0027 \\
Recall    & 0.8141 $\pm$ 0.0026 & 0.7354 $\pm$ 0.0026 \\
F-score   & 0.5247 $\pm$ 0.0025 & 0.7268 $\pm$ 0.0023 \\
\bottomrule
\end{tabular}
}
\caption{Abstract to Claims Task: Mistral-7B}
\label{tab:text2claims}
\end{table}

\begin{table}[H]
\centering
{\small
\begin{tabular}{lrr}
\toprule
Model & Base & Finetuned \\
\midrule
Precision & 0.6222 $\pm$ 0.0038 & 0.7219 $\pm$ 0.0027 \\
Recall    & 0.6364 $\pm$ 0.0034 & 0.7359 $\pm$ 0.0029 \\
F1-score  & 0.5668 $\pm$ 0.0033 & 0.7039 $\pm$ 0.0026 \\
\bottomrule
\end{tabular}
}
\caption{Abstract to Investigation Proposals Task: Mistral-7B}
\label{tab:text2ip}
\end{table}

\section{Examples of Extracted Claims and Investigation Proposals}
\label{appendix:claim-IP-examples}
Tables~\ref{tab:claim-examples} and~\ref{tab:ip-examples} provide a sampling of the extracted claims and investigation proposals. 

\rowcolors{2}{gray!10}{white}
\begin{table*}[h]
\small
\centering
    \begin{tabular}{lp{5cm}p{8cm}}
    \toprule
    \rowcolor{white} 
         \textbf{Award ID} & \textbf{Title} & \textbf{Extracted Claims}\\
         \midrule
         \href{https://www.nsf.gov/awardsearch/showAward?AWD_ID=2324035&HistoricalAwards=false}{2324035} & DMREF: Developing and Harnessing the Platform of Quasi-One-Dimensional Topological Materials for Novel Functionalities and Devices & \begin{minipage}{\linewidth}
\begin{lstlisting}
[
    "Topological insulators are electrically insulating in the bulk but host conductive surface states that are immune to impurities.",
    "Current TI materials face critical challenges that limit their potential.",
    "Quasi-1D structures promise to overcome challenges faced by current TI materials.",
    "Most identified topological insulators are either strongly bonded bulk materials or layered van der Waals materials.",
    ...
]
\end{lstlisting}
\end{minipage} \\
         \href{https://www.nsf.gov/awardsearch/showAward?AWD_ID=9814055&HistoricalAwards=false}{9814055} & Kinks and Surface Potentials & \begin{minipage}{\linewidth}
\begin{lstlisting}
[
    "Atomically flat terraced surfaces for thin TEM samples can be prepared under moderate (10-7 Torr) vacuum conditions by annealing in oxygen or vacuum for materials such as sapphire, SiC and MgO.",
    "\"Forbidden\" Bragg reflections arise from the stacking fault between partial dislocations.",
    "The surface potential is critical for chemical reactions at surfaces, adsorption, catalysis, epitaxy, diffusion bonding process, oxidation, and semiconductor crystal growth.",
    ...
]
\end{lstlisting}
\end{minipage} \\
         \href{https://www.nsf.gov/awardsearch/showAward?AWD_ID=0821136&HistoricalAwards=false}{0821136} & MRI: Acquisition of an Imaging Spherical Aberration Corrector and a Lorentz Lens for Magnetic Materials Characterization & \begin{minipage}{\linewidth}
\begin{lstlisting}
[
    "The attainable spatial resolution of uncorrected Lorentz instruments is in the range 10-15 nm.",
    "Delocalization effects cause significant image blurring in uncorrected Lorentz microscopes.",
    "Recent developments in aberration correction make it possible to correct the spherical aberration of a Lorentz lens.",
    "The size of written bits in state-of-the-art magnetic recording media is comparable to the magnetic resolution of uncorrected Lorentz microscopes.",
    "Transmission electron microscopes have suffered from lens aberration since their invention in the 1930s.",
    "The Hubble space telescope suffered from a similar aberration when first launched.",
    ...
]
\end{lstlisting}
\end{minipage} \\
    \bottomrule
    \end{tabular}
    \caption{A sample of extracted claims from the \DatasetNameMatSci~dataset. Award IDs are hyperlinked to the NSF's Award database.}
    \label{tab:claim-examples}
\end{table*}

\rowcolors{2}{gray!10}{white}
\begin{table*}[h]
\small
\centering
    \begin{tabular}{lp{5cm}p{8cm}}
    \toprule
    \rowcolor{white} 
         \textbf{Award ID} & \textbf{Title} & \textbf{Extracted Investigation Proposals}\\
         \midrule
         \href{https://www.nsf.gov/awardsearch/showAward?AWD_ID=2324035&HistoricalAwards=false}{2324035} & DMREF: Developing and Harnessing the Platform of Quasi-One-Dimensional Topological Materials for Novel Functionalities and Devices & \begin{minipage}{\linewidth}
\begin{lstlisting}
[
    "Predict, design, synthesize, and control topological phases in quasi-1D topological materials.",
    "Design and demonstrate emergent materials, functionalities, and devices, including moir\'e quasi-1D TIs, stable and high temperature quantum spin Hall (QSH) insulators, and quantum intelligent sensors.",
    "Expand research to include other selected quasi-1D materials families through collaborations.",
    "Discover or realize novel topological materials and phases.",
    "Study topological phase transitions and control.",
    ...
]
\end{lstlisting}
\end{minipage} \\
         \href{https://www.nsf.gov/awardsearch/showAward?AWD_ID=9814055&HistoricalAwards=false}{9814055} & Kinks and Surface Potentials & \begin{minipage}{\linewidth}
\begin{lstlisting}
[
    "Observe dislocation kinks by atomic resolution TEM in materials such as sapphire, SiC and MgO.",
    "Use \"forbidden\" Bragg reflections to form lattice images without surface noise.",
    "Determine which process (kink formation, kink migration or obstacles along the dislocation line) limits kink (and hence dislocation) velocity, for given conditions of temperature and stress.",
    "Extend quantitative convergent-beam TEM measurements of bonding in crystals to the RHEED geometry to refine the electrostatic potential extending into the vacuum from ceramic surfaces.",
    "Measure modifications to the surface potential resulting from the deposition of a monolayer or more of atoms.",
    ...
]
\end{lstlisting}
\end{minipage} \\
         \href{https://www.nsf.gov/awardsearch/showAward?AWD_ID=0821136&HistoricalAwards=false}{0821136} & MRI: Acquisition of an Imaging Spherical Aberration Corrector and a Lorentz Lens for Magnetic Materials Characterization & \begin{minipage}{\linewidth}
\begin{lstlisting}
[
    "Acquire an imaging spherical aberration corrector and a Lorentz lens for magnetic materials characterization.",
    "Add these components to an existing FEI Titan 80-300 TEM.",
    "Bring the spatial resolution in Lorentz mode down to less than 1 nm, with negligible delocalization effects.",
    "Enable direct quantitative study of magnetic features at a length scale of around 1 nm.",
    "Obtain new scientific results on material systems for which these observations were previously impossible.",
    "Impact a large number of research groups within CMU, as well as collaborations with local industry and several national laboratories.",
    ...
]
\end{lstlisting}
\end{minipage} \\
    \bottomrule
    \end{tabular}
    \caption{A sample of extracted investigation proposals from the \DatasetNameMatSci~dataset. Award IDs are hyperlinked to the NSF's Award database.}
    \label{tab:ip-examples}
\end{table*}

\section{Examples of Scientific Claim and Investigation Proposal Categories}
\label{appendix:claim-cat-examples}
Please see Table~\ref{tab:claim-cat-examples} and \ref{tab:ip-cat-examples} for the examples.

\begin{table*}[ht]
\small
\resizebox{\textwidth}{!}{%
\rowcolors{3}{white}{white}
\begin{tabular}{p{\textwidth}}
\rowcolor{gray!20}
\textbf{Category: Capability/Application of Technology/Method} \\
Memory-centric computing capitalizes on extensive parallelism in memory arrays. \\
The Illinois group has joined the fixed target COMPASS experiment at CERN. \\
An electronics company is involved in the project, making imaging products in this energy regime. \\
\hline
\rowcolor{gray!20}
\textbf{Category: Definition/Classification} \\
The RV Weatherbird II is owned and operated by the Bermuda Biological Station for Research (BBSR), Inc. \\
The program will include topics such as dark matter, dark energy, inflation, and gravitational waves. \\
The shear zone in question is the Cuyamaca-Laguna Mountains shear zone. \\
\hline
\rowcolor{gray!20}
\textbf{Category: Statement of Problem/Knowledge Gap} \\
Current efforts on analyzing tree-informed compositional data are primarily designed for individual applications. \\
CU began the Guerrero GPS project in 1997. \\
High pressure-low temperature metamorphism is often obscured by post-tectonic thermal equilibration or later deformation and mineral growth. \\
\hline
\rowcolor{gray!20}
\textbf{Category: Experimental Result/Finding/Measurability} \\
Lattice QCD has made important progress. \\
RBP repression is absent when an oncoprotein is present. \\
Over 100 of 650 U.S. electronics fabricators have gone out of business in the past five years, according to a 1999 White Paper by the Interconnection Technology Research Institute. \\
\hline
\rowcolor{gray!20}
\textbf{Category: Established Scientific Fact/Principle} \\
Dynamic programming includes well-known search algorithms like breadth-first search, Dijkstra's algorithm, A*, value iteration and policy iteration for Markov decision processes. \\
The electron carries a magnetic moment. \\
Stars in clusters evolve off the main sequence, become red giants, and ultimately horizontal branch stars. \\
\hline
\rowcolor{gray!20}
\textbf{Category: Observed Phenomenon/Property} \\
The lake level of Laguna Paron was artificially lowered in 1985. \\
Laminated sediments are exposed in Laguna Paron, Peru. \\
The study sites exhibit extreme differences (1 to 2 orders of magnitude) in larval settlement. \\
\hline
\rowcolor{gray!20}
\textbf{Category: Process/Mechanism Description} \\
Exciton-phonon and exciton-exciton interactions contribute to decoherence at finite temperatures. \\
The fidelity of translation is determined by the accuracy of aminoacyl-tRNA selection by ribosomes and synthesis of cognate amino acid/tRNA pairs by aminoacyl-tRNA synthetases. \\
The evaluation process includes both direct and indirect measures of student success and learning. \\
\hline
\rowcolor{gray!20}
\textbf{Category: Hypothesis/Theoretical Prediction} \\
Assemblages that combine human-technology partnerships are stronger than individual humans or machines. \\
Mating advantage in guppies appears to result from female sexual responses to unusual males. \\
The long wavelength part of the CBR spectrum is important for constraining the evolution of the intergalactic medium. \\
\end{tabular}
}
\caption{Scientific claim categories found in \DatasetName~and 3 randomly selected examples for each category.}
\label{tab:claim-cat-examples}
\end{table*}

\begin{table*}[ht]
\small
\resizebox{\textwidth}{!}{%
\rowcolors{3}{white}{white}
\begin{tabular}{p{\textwidth}}
\rowcolor{gray!20}
\textbf{Category: Academic Training and Curriculum Development} \\
Develop a generic geometric interpretation to the wavelet frame transform by studying its relations with differential operators within various variational frameworks. \\
Support participation in the visitor program activities during 2018 - 2020. \\
Measure the contributions of antiquarks to nucleon spin using the PHENIX polarized pp program with an Illinois-led muon trigger upgrade. \\
\hline
\rowcolor{gray!20}
\textbf{Category: Experimental Technique and Tool Development} \\
Develop a generic geometric interpretation to the wavelet frame transform by studying its relations with differential operators within various variational frameworks. \\
Measure the contributions of antiquarks to nucleon spin using the PHENIX polarized pp program with an Illinois-led muon trigger upgrade. \\
Develop a method of creating sulfur ylides with improved yields. \\
\hline
\rowcolor{gray!20}
\textbf{Category: Theoretical Analysis and Computational Modeling} \\
Develop a generic geometric interpretation to the wavelet frame transform by studying its relations with differential operators within various variational frameworks. \\
Deepen understanding about how to recognize the complexity of certain types of computational problems. \\
Support participation in the visitor program activities during 2018 - 2020. \\
\hline
\rowcolor{gray!20}
\textbf{Category: Human/User Study} \\
Focus on the settling and juvenile stages of 7 dominant species within subtidal marine epifaunal communities along the coast of southern New England. \\
Examine the impact of sea ice on the distribution and abundance of zooplankton. \\
Examine and model visual tracking of continuously moving targets in normal human subjects. \\
\hline
\rowcolor{gray!20}
\textbf{Category: Algorithm/Method Development} \\
Develop a generic geometric interpretation to the wavelet frame transform by studying its relations with differential operators within various variational frameworks. \\
Deepen understanding about how to recognize the complexity of certain types of computational problems. \\
Develop a method of creating sulfur ylides with improved yields. \\
\hline
\rowcolor{gray!20}
\textbf{Category: Policy/Guidelines/Standards Work} \\
Design, fabricate, assemble, align, test, integrate, and calibrate a sensitive CCD camera system. \\
Provide funding to offset registration fees for about 12 graduate students or postdocs at the COSMO-16 conference. \\
Replace two semi-conductor detectors in the Neutron Activation Laboratory. \\
\hline
\rowcolor{gray!20}
\textbf{Category: Interpretability/Alignment Analysis} \\
Understand and correct for hidden assumptions in Bayesian inference algorithms. \\
Develop assemblages for human-technology partnerships in visually based cognition-oriented tasks in radiology. \\
Systematically investigate and proactively prevent specious configurations. \\
\hline
\rowcolor{gray!20}
\textbf{Category: Deployment/Field Study} \\
Develop a method of creating sulfur ylides with improved yields. \\
Design, fabricate, assemble, align, test, integrate, and calibrate a sensitive CCD camera system. \\
Measure chlorofluorocarbons (CFC-11, CFC-12, CFC-113) on the 26 degrees N transect in winter 2004. \\
\hline
\rowcolor{gray!20}
\textbf{Category: Tooling/Systems/Infrastructure} \\
Deepen understanding about how to recognize the complexity of certain types of computational problems. \\
Design, fabricate, assemble, align, test, integrate, and calibrate a sensitive CCD camera system. \\
Understand and correct for hidden assumptions in Bayesian inference algorithms. \\
\hline
\rowcolor{gray!20}
\textbf{Category: Empirical Benchmarking/Evaluation} \\
Measure the contributions of antiquarks to nucleon spin using the PHENIX polarized pp program with an Illinois-led muon trigger upgrade. \\
Obtain accurate colors and brightnesses of the brighter stars in 50 globular clusters over a two-year period. \\
Develop a new density cumulant functional theory. \\
\hline
\end{tabular}
}
\caption{Investigation proposal categories found in \DatasetName~and 3 examples for each category.}
\label{tab:ip-cat-examples}
\end{table*}

\section{Error Analysis Examples}
\label{app:error-examples}

\subsection{Claims}

Of the three proposed tasks, we consider the claim extraction task as a canonical task for performing error analysis. To do so, we consider another 120 awards from the test portion of \DatasetNameMatSci~and \DatasetNameTwentyK. These were stratified samples across the five areas of interest (similar to Section ~\ref{sec:eval-claude-extraction}). We then generate the claims using a Mistral-7B model finetuned on \DatasetNameTwentyK, resulting in 802 claims. A careful examination revealed around 2.6\% of the generated claims were incorrect. To dive deeper, we categorized the erroneous claims into 5 categories. We list them here with examples:

\paragraph{1. Overconfidence: }
The claim can be overconfident about information that has qualifiers in the supporting document text (award abstract).

{\small
\begin{tcolorbox}[title={\textbf{Award ID:} 9820570}]
\textbf{Extracted Claim:} The research areas include knot theory, immiscible fluids and geodesic nets, ergodic theory, commutative algebra and vector-valued forms.\\
\textbf{Analysis:} The abstract states 'probably in the areas of,' indicating potential areas, not certainty.
\end{tcolorbox}
}

\paragraph{2. Mixing Information:}
The claim can mix information from two sentences together to form wrong information.

{\small
\begin{tcolorbox}[title={\textbf{Award ID:} 1205671}]
\textbf{Extracted Claim:} The SEAQUEST experiment at Fermilab has successfully measured the asymmetry of up and down anti-quarks in the nucleon.\\
\textbf{Analysis:} The abstract mentions that SEAQUEST will follow the successful E866 measurement with more precise data, and thus it does not say SEAQUEST has already successfully measured that, but the success is describing the previous E866.
\end{tcolorbox}
}

\paragraph{3. Overgeneralization:}
The claim can overgeneralize what the supporting document implies.

{\small
\begin{tcolorbox}[title={\textbf{Award ID:} 0957482}]
\textbf{Extracted Claim:} The methodology is potentially environmentally benign.\\
\textbf{Analysis:} The abstract mentions non-dangerous chemicals but does not specifically state that the methodology is environmentally benign.
\end{tcolorbox}
}

\paragraph{4. Information Omission:}
The claim might omit important information from the abstract and thus the meaning is changed.

{\small
\begin{tcolorbox}[title={\textbf{Award ID:} 9409461}]
\textbf{Extracted Claim:} Frequency-domain techniques can display trade-offs between output performance and sensitivity reduction.\\
\textbf{Analysis:} The claim frames output performance and sensitivity reduction as two separate quantities and leaves out bandwidth, so it does not accurately reflect the abstract.
\end{tcolorbox}
}

\paragraph{5. Hallucinations about Administrative Metadata:}
The model can sometimes hallucinate claims regarding where the funding is from and which institutions are included. While hallucination is a serious issue, it is worth noting that for this dataset and model scientific claims seem to be rarely hallucinated. In our study, all hallucinations were connected with administrative metadata.

\begin{tcolorbox}[title={\textbf{Award ID:} 0542751}]
\textbf{Claim:} The award is funded under the American Recovery and Reinvestment Act of 2009 (Public Law 111-5).\\
\textbf{Reasoning:} This claim is not mentioned in the abstract.
\end{tcolorbox}

To mitigate these errors, uncertainty calibration and prompting strategies can reduce overconfidence and overgeneralization, encouraging the model to reflect source qualifiers. Fine-tuning with more annotated data and enforcing stricter alignment between claims and source text can address mixing information and omission issues. Retrieval-augmented generation and chain-of-thought prompting may also promote better grounding. For hallucinations about administrative metadata, entity verification or output constraints based on structured data can help. Combining these approaches with human-in-the-loop evaluation might further improve claim extraction reliability.

We performed a similar error analysis on claims extracted from Claude (See section ~\ref{sec:claude-claim-extraction}). Our findings revealed a smaller error-rate (2.1\% as opposed to 2.6\%), and of the only 10 erroneous claims, 5 were hallucinations of administrative data.

\subsection{Investigation Proposals}
We additionally performed an error analysis on investigation proposals, following the same procedure as for claims
Among 120 awards from the test portion of \DatasetNameMatSci{} and \DatasetNameTwentyK{}, and generated the investigation proposals using a Mistral-7B model finetuned on \DatasetNameTwentyK{}, resulting in 833 proposals. A careful examination revealed around 2.4\% of the generated investigation proposals were incorrect.
To dive deeper, we categorized the erroneous proposals into 4 categories.
We list them here with examples:

\paragraph{No investigation proposals.}
The abstract itself does not have investigation proposals and the model forcefully generates some that are not proposals.

{\small
\begin{tcolorbox}[title={\textbf{Award ID:} 1642020}]
\textbf{Investigation Proposal:} The conference aims to advance understanding in cosmology through the presentation and discussion of new research findings.\\
\textbf{Analysis:} It implies discussion and presentation rather than a forward-looking research goal, and the abstract only seeks funding to cover junior students’ registration fees, so it contains no actual investigation proposals.
\end{tcolorbox}
}

\paragraph{Content Mismatch Error.}
The investigation proposal does not accurately reflect the information in the abstract—either by introducing concepts not mentioned, omitting key elements, or misrepresenting the scope or focus of the abstract’s content.

{\small
\begin{tcolorbox}[title={Experimental Technique and Tool Development: Develop new experimental tools and techniques for data collection or experimentation.}]
\textbf{Investigation Proposal:} Access and design different novel nano-motion components.\\
\textbf{Reasoning:} The abstract focuses on designing and characterizing catalytic nanomotors with different dynamic behaviors, but it does not mention accessing or broadly designing novel nano-motion components.

\end{tcolorbox}
}

\paragraph{Overspecification.} The proposal extracted is more specific than what is actually mentioned in the abstract, containing non-existing details.

{\small
\begin{tcolorbox}[title={Experimental Technique and Tool Development: Develop new experimental tools and techniques for data collection or experimentation.}]
\textbf{Investigation Proposal:} Design and integrate novel nanomachines.\\
\textbf{Reasoning:} The abstract focuses on designing and characterizing nanomotors, not on integrating nanomachines. ``Integrated nanomachinary systems'' is mentioned only as a motivation, not as a research activity.

\end{tcolorbox}
}

\paragraph{Existing Work.} The claim is about existing work instead of a forward-looking statement.

{\small
\begin{tcolorbox}[title={Algorithm/Method Development: Propose novel algorithms or procedures for solving specific problems or improving performance.}]
\textbf{Investigation Proposal:} Apply the developed approach to find subtle errors in non-trivial designs.\\
\textbf{Reasoning:} The proposal describes a planned research activity, whereas the abstract only refers to a past application, not a proposed investigation.

\end{tcolorbox}
}

\section{Potential Risks}
\DatasetName{} opens new opportunities for large-scale scientific text analysis, but responsible use is important. As the dataset is automatically constructed, some extraction errors or omissions may remain, underscoring the need for careful validation in downstream applications. Its coverage of NSF award abstracts may reflect domain-specific language and institutional styles, which can inform analyses but may also introduce biases if not accounted for. Finally, while the dataset enables powerful new capabilities, users should ensure appropriate use to avoid generating or disseminating unverified claims.

\section{AI Writing/Coding Assistance Disclosure}
In accordance with the ACL Policy on AI Writing Assistance\footnote{\url{https://www.aclweb.org/adminwiki/index.php/ACL_Policy_on_Publication_Ethics\#Guidelines_for_Generative_Assistance_in_Authorship}}, the authors attest that we used generative AI tools for assistance purely with the language of the paper, including spell checking, grammar fixes, and proof reading. Additionally, we used GPT-4o to fix LaTeX issues, and to generate LaTeX tables from spreadsheets. In all such uses, the outputs were verified by the first author for correctness.

\end{document}